\documentclass{wscpaperproc}
\usepackage{latexsym}
\usepackage{graphicx}
\usepackage{mathptmx}
\usepackage{newtxtext} 
\usepackage{newtxmath} 
\usepackage[T1]{fontenc}

\usepackage{amsmath}
\usepackage{amsfonts}
\usepackage{booktabs}

\usepackage[pdftex,colorlinks=true,urlcolor=blue,citecolor=black,anchorcolor=black,linkcolor=black]{hyperref}

\begin{document}

%
%

\pagestyle{fancyplain}

\thispagestyle{plain}
\firstPageHead{}

\chead{\fancyplain{}{\itshape Wang, Ustun, and McGroarty}}

\rhead{}
\cfoot{}
\renewcommand{\headrulewidth}{0pt} 

\makeatletter
\let\@internalcite\cite
\def\cite{\def\@citeseppen{-1000}%
    \def\@cite##1##2{(##1\if@tempswa , ##2\fi)}%
    \def\citeauthoryear##1##2##3{##1 ##3}\@internalcite}
\def\citeNP{\def\@citeseppen{-1000}%
    \def\@cite##1##2{##1\if@tempswa , ##2\fi}%
    \def\citeauthoryear##1##2##3{##1 ##3}\@internalcite}
\def\citeN{\def\@citeseppen{-1000}%
    \def\@cite##1##2{##1\if@tempswa, ##2)\else{}\fi}%
    \def\citeauthoryear##1##2##3{##1 (##3)}\@citedata}
\def\citeA{\def\@citeseppen{-1000}%
    \def\@cite##1##2{(##1\if@tempswa , ##2\fi)}%
    \def\citeauthoryear##1##2##3{##1}\@internalcite}
\def\citeANP{\def\@citeseppen{-1000}%
    \def\@cite##1##2{##1\if@tempswa , ##2\fi}%
    \def\citeauthoryear##1##2##3{##1}\@internalcite}
\def\shortcite{\def\@citeseppen{-1000}%
    \def\@cite##1##2{(##1\if@tempswa , ##2\fi)}%
    \def\citeauthoryear##1##2##3{##2 ##3}\@internalcite}
\def\shortciteNP{\def\@citeseppen{-1000}%
    \def\@cite##1##2{##1\if@tempswa , ##2\fi}%
    \def\citeauthoryear##1##2##3{##2 ##3}\@internalcite}
\def\shortciteN{\def\@citeseppen{-1000}%
    \def\@cite##1##2{##1\if@tempswa, ##2\else{}\fi}%
    \def\citeauthoryear##1##2##3{##2 (##3)}\@citedata}
\def\shortciteA{\def\@citeseppen{-1000}%
    \def\@cite##1##2{(##1\if@tempswa , ##2\fi)}%
    \def\citeauthoryear##1##2##3{##2}\@internalcite}
\def\shortciteANP{\def\@citeseppen{-1000}%
    \def\@cite##1##2{##1\if@tempswa , ##2\fi}%
    \def\citeauthoryear##1##2##3{##2}\@internalcite}
\def\citeyear{\def\@citeseppen{-1000}%
    \def\@cite##1##2{(##1\if@tempswa , ##2\fi)}%
    \def\citeauthoryear##1##2##3{##3}\@citedata}
\def\citeyearNP{\def\@citeseppen{-1000}%
    \def\@cite##1##2{##1\if@tempswa , ##2\fi}%
    \def\citeauthoryear##1##2##3{##3}\@citedata}
%
%
%
\def\@citedata{%
    \@ifnextchar [{\@tempswatrue\@citedatax}%
                  {\@tempswafalse\@citedatax[]}%
}

\def\@citedatax[#1]#2{%
\if@filesw\immediate\write\@auxout{\string\citation{#2}}\fi%
  \def\@citea{}\@cite{\@for\@citeb:=#2\do%
    {\@citea\def\@citea{, }\@ifundefined
       {b@\@citeb}{{\bf ?}%
       \@warning{Citation `\@citeb' on page \thepage \space undefined}}%
{\csname b@\@citeb\endcsname}}}{#1}}%

%
\def\@citex[#1]#2{%
\if@filesw\immediate\write\@auxout{\string\citation{#2}}\fi%
  \def\@citea{}\@cite{\@for\@citeb:=#2\do%
    {\@citea\def\@citea{; }\@ifundefined
       {b@\@citeb}{{\bf ?}%
       \@warning{Citation `\@citeb' on page \thepage \space undefined}}%
{\csname b@\@citeb\endcsname}}}{#1}}%

%
\def\@biblabel#1{}
\makeatother



\newdimen\bibindent
\bibindent=0.0em
\def\thebibliography#1{\section*{\refname}\list
   {}{\settowidth\labelwidth{[#1]}
   \leftmargin\parindent
   \itemindent -\parindent
   \listparindent \itemindent
   \itemsep 0pt
   \parsep 0pt}
   \def\newblock{}
   \sloppy
   \sfcode`\.=1000\relax}


\setlength{\baselineskip}{12.7pt}

\title{\textsc{A data-driven discretized CS:GO simulation environment to facilitate strategic multi-agent planning research}}

\author{\begin{center}Yunzhe Wang\textsuperscript{1,2}, Volkan Ustun\textsuperscript{1}, and Chris McGroarty\textsuperscript{3}\\
[11pt]
\textsuperscript{1}USC Institute for Creative Technologies, Los Angeles, CA, USA\\
\textsuperscript{2}Dept.~of Computer Science, University of Southern California, Los Angeles, CA, USA\\
\textsuperscript{3}U.S. Army Combat Capabilities Development Command, Orlando, FL, USA\end{center}
}

\maketitle
\vspace{-12pt}

\section*{ABSTRACT}
Modern simulation environments for complex multi-agent interactions must balance high-fidelity detail with computational efficiency. We present DECOY, a novel multi-agent simulator that abstracts strategic, long-horizon planning in 3D terrains into high-level discretized simulation while preserving low-level environmental fidelity. Using Counter-Strike: Global Offensive (CS:GO) as a testbed, our framework accurately simulates gameplay using only movement decisions as tactical positioning—without explicitly modeling low-level mechanics such as aiming and shooting. Central to our approach is a waypoint system that simplifies and discretizes continuous states and actions, paired with neural predictive and generative models trained on real CS:GO tournament data to reconstruct event outcomes. Extensive evaluations show that replays generated from human data in DECOY closely match those observed in the original game. Our publicly available simulation environment provides a valuable tool for advancing research in strategic multi-agent planning and behavior generation.
\section{Introduction}
\label{sec:intro}

Team-based multiplayer strategy games have emerged as grand challenge domains for multi-agent learning and long-horizon planning. Breakthroughs in complex games such as StarCraft II and Dota 2—where AI agents have achieved human-expert or even superhuman performance through large-scale self-play training \shortcite{baker2019emergent,vinyals2019grandmaster,berner2019dota,team2021open}—demonstrate that, given sufficient simulation and training, sophisticated strategies can be discovered even in environments characterized by long time horizons, imperfect information, and high-dimensional state-action spaces. However, these successes come at the expense of enormous computational costs, and a prevailing trend in the field is to scale performance by increasing model size and training on larger datasets with more simulation steps \shortcite{neumann2022scaling,obando2024mixtures,kaplan2020scaling}. In numerous real-world settings—team sports, emergency response, and search and rescue among them—high-fidelity simulators are either unavailable or prohibitively expensive in both computational and financial terms. Furthermore, many existing strategy game environments, like StarCraft and Dota, utilize isometric or ``pseudo-3D'' perspectives rather than authentic three-dimensional worlds with first-person viewpoints that more accurately represent real-world scenarios.

Recent advances in offline, model-based, and in-context (reinforcement) learning—such as world models—offer promising paths to address these challenges. Offline RL improves sample efficiency by learning from fixed datasets, while mitigating issues like distributional shift and overestimation bias, and has shown success in both game and robotics domains \shortcite{reed2022generalist,mathieu2023alphastar,o2024open,laskin2022context,nikulin2024xland}. Similarly, world models have enabled effective planning in compact latent spaces, achieving strong performance in single-agent tasks \shortcite{ha2018world,hafner2023mastering,li2024open,garrido2024learning}. Yet, their potential for team-based strategic decision-making scenarios remains relatively unexplored.

In parallel, real-world sports analytics—especially in soccer—has driven the development of tactical AI models that learn to predict and generate team behaviors and strategies from player trajectories and game events \shortcite{tuyls2021game,omidshafiei2022multiagent,wang2024tacticai}. However, sports data often remains inaccessible due to privacy concerns and proprietary restrictions \shortcite{socolow2017game}. In contrast, the electronic sports (e-sports) domain provides abundant, high-fidelity data that is more amenable to large-scale modeling and experimentation \shortcite{xenopoulos2022esta,xenopoulos2023data}. Counter-Strike: Global Offensive (CS:GO), a popular 5v5 first-person shooter, offers such rich strategic complexity—encompassing coordinated team tactics, resource management, and decision-making under partial observability. In each match, two teams—the Terrorists (T) and Counter-Terrorists (CT)—compete in a series of objective-based rounds, where the Terrorists aim to plant a bomb at a designated site, and the Counter-Terrorists must prevent the plant or defuse the bomb once planted. Success depends not only on mechanical skill but also on timing, positioning, communication, and resource management. Recent work has demonstrated that agents trained directly on human gameplay logs in CS:GO can replicate expert-level movement \shortcite{durst2024csgo}. However, a fast, learning-compatible simulator for the game remains unavailable, despite its potential for enabling sample-efficient training and large-scale evaluation of AI policies in high-dimensional, multi-agent strategic decision-making settings.

In this paper, we introduce \textsc{DECOY}, a novel simulation environment designed to support research in strategic multi-agent planning using CS:GO as a testbed. DECOY aligns high-fidelity human gameplay data into an abstracted, discretized simulation framework that enables efficient modeling of complex team behaviors and long-horizon tactical decisions. Without explicitly modeling low-level mechanics such as aiming, recoil, or animation, DECOY focuses on strategic-level decision-making. It leverages a waypoint-based navigation system for action abstraction, along with predictive and generative models trained with real CS:GO tournament data to estimate action outcomes such as shooting damage and engagement results. This abstraction enables fast simulation and reduced action complexity while preserving the original game data distribution, offering a practical and scalable platform for advancing multi-agent learning and strategic behavior generation in e-sports and beyond. Our simulator and pre-trained models are publicly available at \href{https://github.com/HATS-ICT/decoy}{github.com/HATS-ICT/decoy}.

\section{The DECOY Simulator}

\begin{figure}[!htbp]
\centering
\includegraphics[width=0.95\linewidth]{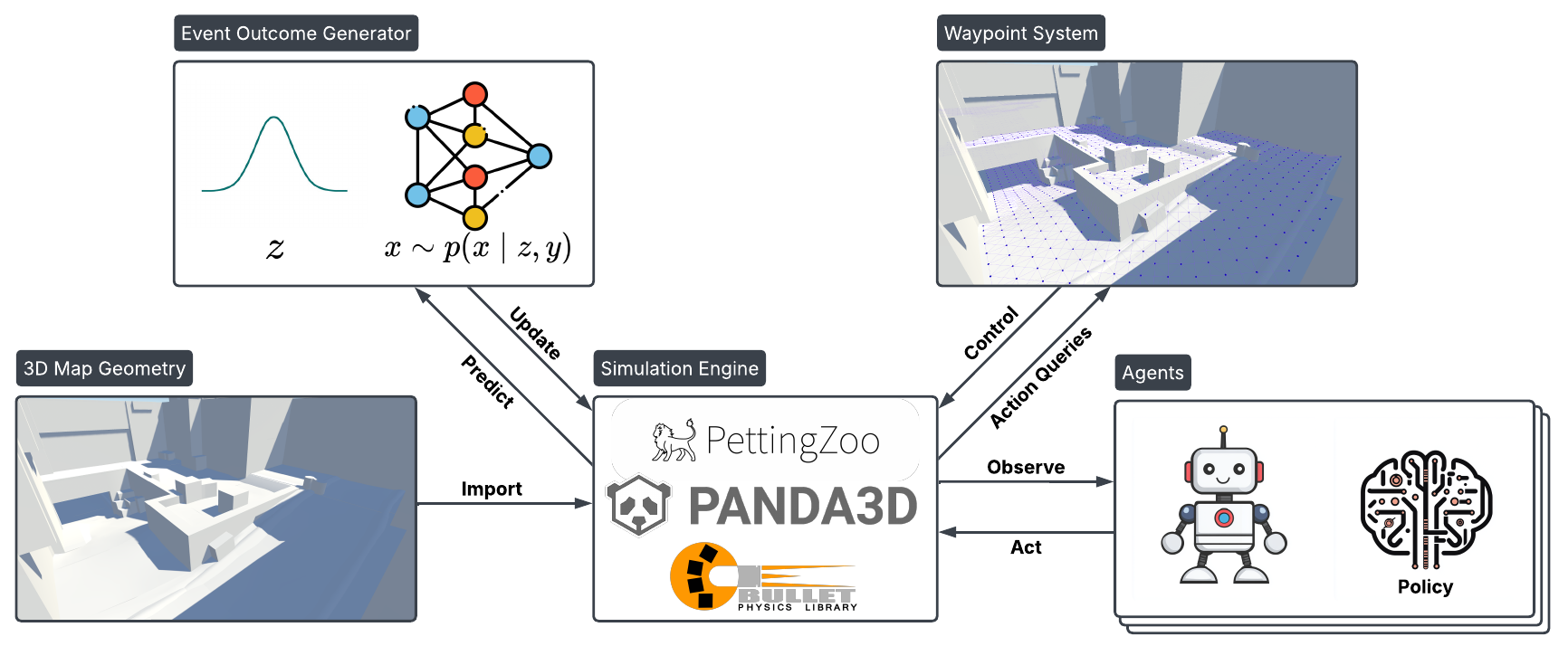}
\caption{Overview of the DECOY simulation framework. A 3D CS:GO map, Dust II (de\_dust2) is imported into the simulation engine and discretized by generated waypoints. Given agent game states, pre-trained predictive and generative models reconstruct outcomes of low-level interactions, such as shooting and damage.}
\label{fig:framework_diagram}
\end{figure}

We present \textbf{DECOY} (\textbf{D}iscrete-\textbf{E}vent \textbf{CO}unter-Strike simul\textbf{Y}tor), a Python-based, multi-agent learning simulation environment to support strategic planning in 3D terrain in the game CS:GO. DECOY provides a Gym-style Application Programming Interface (API) to benchmark standard Multi-Agent Reinforcement Learning (MARL) algorithms and is calibrated with human-trajectory tournament data for offline and model-based learning. The framework (Fig \ref{fig:framework_diagram}) comprises three core components: (1) a physics-based 3D environment, (2) a waypoint-based discretization system for movement abstraction, and (3) neural predictive and generative models for event outcome prediction and generation. We introduce the framework in this section and evaluate its performance and alignment with the original game in section \ref{sec:results}

\subsection{3D Environment}
The simulation environment is built with Panda3D \shortcite{panda3d} and the Bullet Physics library for agent control and collision detection. Environment stepping, game states, and actions follow the PettingZoo API \shortcite{terry2021pettingzoo}, ensuring compatibility with standard MARL algorithms. While CS:GO is typically 5v5, our setup supports learning between any number of agents.

To replicate accurate 3D map and physics, we decompiled the \texttt{de\_dust2} map using online tool BSPSource \shortcite{bspsource}, retaining only geometry and removing decorative assets. Agent physics—dimensions, speed, and jumping—follow CS:GO specifications, with Hammer Units from Source Engine converted to meters. The simulation proceeds as discrete events with incremental time steps, with actions guided by a waypoint structure (Section~\ref{sec:waypoint}). The state space includes data for 10 agents—position, view angle, weapon, armor, helmet—and the bomb’s position/state. The action space covers movement, view direction, and stopping.

\subsection{Waypoint Discretization}
\label{sec:waypoint}

We adopted the same waypoint control setup as \shortciteN{koresh2024improving}, where prior work has demonstrated its effectiveness in aligning movement trajectories \shortcite{ustun2024geospecific}. The waypoint graph is a directed graph where vertices denote specific 3D location, and edges represent feasible direct movements between positions. The graph forms a lattice over the map, with fixed distances between waypoints. Each edge is labeled with one of the eight compass directions (N, NE, E, SE, S, SW, W, NW). 

Because map navigation involves elevation changes (e.g., ramps), jumping onto boxes, and the fact that diagonal directions cover more distance than cardinal ones, agents may require different numbers of frames to complete the eight directional actions. To address this, agents operate in an on-demand fashion: a new action is requested after the agent reaches its current waypoint, and agent actions proceed asynchronously.

We extended the waypoint generation algorithm of \shortciteN{koresh2024improving} to handle complex 3D environments with varying elevation and multi-floor layouts using a three-stage process:

\begin{itemize}
    \item \textbf{Breadth-First Search (BFS) Generation:} Starting from manually specified coordinates, we use a BFS style algorithm to lay waypoints as a bidirectional graph across the 3D map.  Various ray tests find neighboring points at proper elevations and avoid placing waypoints inside walls.
    
    \item \textbf{Drop-Floor Check:} In a second pass, we iterate over all generated waypoints and add singly directed edges from higher to lower elevations that were not previously connected. This enables agents to drop from higher floors to lower ones.
    
    \item \textbf{Waypoint Verification:} In the final pass, we iterate over all directed edges by placing actual agents to simulate movement between endpoints. This process eliminates edge cases, such as obstacles that block agent movement but were missed in earlier checks. Any blocked edges are removed. The resulting waypoint graph is guaranteed to have no dead ends and to be strongly connected.
\end{itemize}

In rare cases where critical areas are not covered by the waypoint generation algorithm (e.g., the narrow jumping edge on the X-Box), we manually place waypoints.

\subsection{Human Trajectory Dataset}

Although the environment is abstracted and discretized, we calibrate its dynamics and data distribution using real human gameplay data. This is achieved by training predictive and generative models (see Section~\ref{sec:damage_model}) and evaluating environment fidelity through replaying original human gameplay trajectories (see Section~\ref{sec:results}). Specifically, we use the Esports Trajectories \& Actions (ESTA) dataset \shortcite{xenopoulos2022esta}, a comprehensive collection of professional CS:GO match data. This dataset captures complete player trajectories throughout game rounds, including player states, shooting events, damage instances, bomb status, and game outcomes.

We preprocess the dataset by treating each round independently and extracting relevant information for both replay and model training. The extraction process involves parsing official match recordings from professional tournaments to build a structured representation of player movements and interactions.

For each player $i \in \{1, 2, \dots, 10\}$ at time step $t$, we extract a state vector
\[
s_i^t = (p_i^t, v_i^t, h_i^t, e_i^t)
\]
where $p_i^t = (x_i^t, y_i^t, z_i^t) \in \mathbb{R}^3$ is the position, $v_i^t \in [0, 360)$ is the view angle, $h_i^t \in [0, 100]$ is the health, and $e_i^t$ encodes the equipment status (including weapon type, armor value, and helmet presence).

A player’s trajectory over a complete round is represented as a sequence $\tau_i = (s_i^1, s_i^2, \dots, s_i^T)$, where $T$ is the total number of time steps in the round. These trajectories are synchronized to a common timeline, allowing us to reconstruct the complete game state at any moment as $S^t = (s_1^t, s_2^t, \dots, s_{10}^t)$.

We also extract all damage events as tuples $(a, v, d, g, t)$, where $a$ is the attacker index, $v$ is the victim index, $d \in \mathbb{R}^+$ is the damage amount, $g \in \mathcal{G}$ is the hit group $\mathcal{G} = \{\text{Head}, \text{Neck}, \text{Chest}, \text{Stomach}, \text{Arm}, \text{Leg}\}$, and $t$ is the timestamp. Since multiple shots may occur between movement updates, we align damage events with the nearest movement trajectory time tick and aggregate damage by summing over aligned events.

The processed dataset contains approximately 1.5k matches, comprising over 41k rounds and 8 million frames across 8 maps. Each round contains up to 155 seconds of gameplay, sampled at 2 Frames Per Second (FPS). We split the dataset into training, validation, and test sets using an 8:1:1 ratio. All available data is used for training, while inference and replay evaluations are conducted exclusively on the Dust II map.

\subsection{Damage Event Models}
\label{sec:damage_model}

A key challenge in our simulation environment is accurately modeling combat outcomes without requiring agents to control precise aiming mechanics. To address this, we developed two probabilistic models that predict engagement outcomes: (1) a \textbf{Damage Indicator Predictor (DIP)}, which determines whether any damage is likely to occur in a given tactical situation; and (2) a \textbf{Damage Outcome Generator (DOG)}, which if damages is predicted, produces a detailed combat outcomes such as damage amount, death events, and hit locations, based on the learned data distribution.

For simplicity, we model damage events as interactions between agent pairs—specifically, one attacker and one victim. During each inference step, a batch of $5 \times 5 \times 2 = 50$ agent pairs (covering all possible attacker-victim combinations across both teams and directions) is processed to assess the complete tactical situation.

We describe the model architectures for DIP and DOG in Section~\ref{sec:model_dip} and Section~\ref{sec:model_dog}, respectively, and present model evaluation results in Section~\ref{sec:model_eval}.

\subsubsection{Damage Indicator Predictor (DIP)}
\label{sec:model_dip}

The Damage Indicator Predictor (DIP) is a binary classifier that predicts whether a damage event will occur between an attacker and a victim based on contextual game state features. At time step $t$, input features are $x = \{s^t_{\text{attacker}},\ s^t_{\text{victim}},\ x_{\text{mapId}}\}$, including map ID, positions, angles, weapon, and armor status. These are encoded by a shared Game State Encoder $f_{\text{enc}}(\cdot)$, a multi-layer perceptron (MLP), into a latent representation:
\[
\mathbf{h}_{\text{cond}} = f_{\text{enc}}(x) \in \mathbb{R}^{d}
\]
The encoded features are passed to a classification head:
\[
\hat{y}_{\text{DIP}} = f_{\text{cls}}(\mathbf{h}_{\text{cond}}) \in \mathbb{R}
\]
A sigmoid activation is applied during training for binary cross-entropy loss, and thresholded during inference. DIP evaluation is presented in Section~\ref{sec:model_eval_dip}.

\subsubsection{Damage Outcome Generator (DOG)} 
\label{sec:model_dog}

If DIP predicts a potential damage event, the input is passed to DOG—a Conditional Variational Auto-Encoder (CVAE)~\shortcite{kingma2013auto,sohn2015learning}—which generates damage amounts and hit group.

The rationale for using a generative model for damage outcomes, rather than a discriminative one, stems from the nature of damage mechanics in CS:GO. Damage is influenced by multiple factors, most prominently weapon type and hit group. For instance, a headshot with an Avtomat Kalashnikova (AK)-47 typically results in a one-shot kill, while a body shot does not. However, since our simulator does not explicitly implement aiming—and it is non-trivial to implement realistic aiming without producing behavior that resembles an aimbot—hit locations are not directly modeled.

We experimented with a discriminative model that directly predicts damage values, as well as a multitask variant that predicts both damage and hit group using two output heads. Both approaches failed to capture the original data distribution: the former tends to regress toward the statistical mean, while the latter captures only marginal distributions and fails to model the joint dependency between hit group and damage. 

Let $\mathbf{h}_{\text{cond}}$ denote the conditional features from the game state encoder, $\mathbf{d} \in \mathbb{R}$ be the ground-truth damage value, and $\mathbf{g} \in \{0,1\}^{k}$ be the one-hot hit group label, where $k$ is the number of hit groups.

\textit{Training Phase.} During training, the encoder maps inputs to latent parameters:

\[
\mathbf{z}_{\text{in}} = \text{concat}(f_{\text{embed}}(\mathbf{d}), f_{\text{embed}}(\mathbf{g}), \mathbf{h}_{\text{cond}}) \in \mathbb{R}^{m}
\]
\[
\mu, \log\sigma^2 = f_{\text{enc}}^{\text{VAE}}(\mathbf{z}_{\text{in}})
\]

A latent variable $\mathbf{z}$ is sampled using the reparameterization trick:

\[
\mathbf{z} = \mu + \sigma \odot \epsilon, \quad \epsilon \sim \mathcal{N}(0, \mathbf{I})
\]

The sampled latent vector and conditional features are passed to the decoder:

\[
\mathbf{z}_{\text{dec}} = \text{concat}(\mathbf{z}, \mathbf{h}_{\text{cond}})
\]
\[
\hat{\mathbf{d}}, \hat{\mathbf{g}} = f_{\text{dec}}^{\text{VAE}}(\mathbf{z}_{\text{dec}})
\]

Here, $\hat{\mathbf{d}}$ is the predicted damage value and $\hat{\mathbf{g}} \in \mathbb{R}^{k}$ are the logits for the hit group classification.

The model is trained using a loss function derived from the Evidence Lower BOund (ELBO), with both supervised reconstruction terms and a Kullback-Leibler (KL) divergence regularization:

\[
\mathcal{L} = \mathbb{E}_{q(\mathbf{z} \mid \mathbf{d}, \mathbf{g})} \left[ 
\lambda_d \cdot \|\hat{\mathbf{d}} - \mathbf{d}\|^2 
+ \lambda_g \cdot \text{CE}(\hat{\mathbf{g}}, \mathbf{g}) 
\right] 
+ \lambda_{\text{KL}} \cdot D_{\text{KL}}\left(q(\mathbf{z} \mid \mathbf{d}, \mathbf{g}) \,\|\, p(\mathbf{z})\right)
\]

where $\text{CE}(\cdot)$ denotes the cross-entropy loss over hit groups, and $\lambda_{\text{d}}, \lambda_{\text{g}}, \lambda_{\text{KL}}$ are tunable hyperparameters that control the relative weighting between the regression term, classification loss, and KL regularization.

\textit{Inference Phase.} During generation, the VAE encoder is removed, and the latent variable $\mathbf{z} \sim \mathcal{N}(0, \mathbf{I})$ is sampled directly from the prior and combined with the conditional context to decode sampled damage values and hit group predictions.

We evaluate both the reconstruction mode and the generative mode (sampling from the prior) in Section~\ref{sec:model_eval_dog}, and analyze the learned latent space $\mathbf{z}$ in Section~\ref{sec:latent}.

\begin{figure}[!htbp]
\centering
\includegraphics[width=0.8\linewidth]{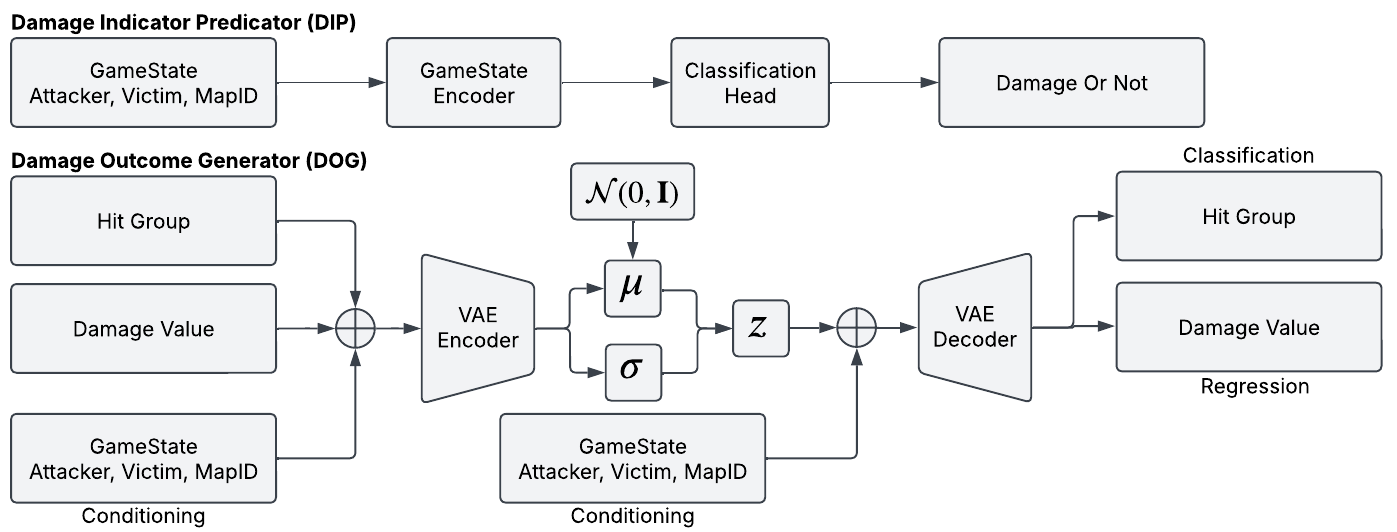}
\caption{Architecture of the Damage Indicator Predictor (DIP) and Damage Outcome Generator (DOG). A shared Game State Encoder processes context, while a Variational Auto-Encoder (VAE) maps damage and hit group outcomes to a latent space. During generation, only the decoder is used by sampling latent variables from $\mathcal{N}(0, I)$.}
\label{fig:damage_model}
\end{figure}

\section{Results}
\label{sec:results}

We evaluate DECOY on three key aspects: (1) its simulation speed, (2) the ability of the damage event models to replicate the original data distribution, and (3) its accuracy in replaying the original game in terms of player movements, damage outcomes, and final game results.

\subsection{Simulation Speed}

To evaluate the efficiency of the simulator, we conducted a timing benchmark by running the simulation with varying numbers of agents. Each configuration was run until a total of 10,000 agent decision steps had been completed. Graphics rendering was disabled, and the physics engine was stepped at 60 ticks per second to match the original game's physics fidelity.

The results are summarized in Table~\ref{tab:sim_performance}. Our simulator achieved approximately a 16x speed-up (966 Ticks/sec) over real-time performance in the standard 5v5 (10-agent) scenario. However, a noticeable slowdown was observed as the number of agents increased, due to the increased computational load in the physics simulation.

\begin{table}[h]
\centering
\caption{Simulation speed under varying agent counts.}
\label{tab:sim_performance}
\resizebox{0.9\linewidth}{!}{%
\begin{tabular}{ccccccc}
\toprule
\textbf{\# Agents} & \textbf{\# Physics Ticks} & \textbf{\# Decisions} & \textbf{Wall Time (s)} & \textbf{Physics Time (s)} & \textbf{Ticks/sec} & \textbf{Time Scale} \\
\midrule
2   & 47,373 & 10,000  & 10.96 & 789.55 & 4322.35 & 72.03 \\
6   & 15,897 & 10,000  & 10.07 & 264.95 & 1578.56 & 26.32 \\
10  & 9,472  & 10,000  & 9.80  & 157.87 & 966.53  & 16.11 \\
20  & 4,742  & 10,000  & 9.73  & 79.03  & 487.25  & 8.12  \\
100 & 944    & 10,000  & 9.48  & 15.73  & 99.58   & 1.66  \\
\bottomrule
\end{tabular}%
}
\end{table}

\subsection{Model Evaluation}
\label{sec:model_eval}

We evaluate the damage models in terms of classification accuracy, generative fidelity, and representational structure. Specifically, we assess: (1) the ability of the Damage Indicator Predictor (DIP) to detect whether a damage event occurs; (2) the Damage Outcome Generator's (DOG) capability to reconstruct and generate realistic damage values and hit group distributions; and (3) the structure of the learned latent space in the DOG model.

\subsubsection{Damage Indicator Model Evaluation}
\label{sec:model_eval_dip}


The Damage Indicator Predictor (DIP) achieved an accuracy of 90.8\%, an F1-score of 0.913, precision of 87.7\%, recall of 94.8\%, an Area Under the Receiver Operating Characteristic Curve (AUC-ROC) of 0.951, and an Average Precision (AP) of 0.927 on the holdout test dataset. The high accuracy and F1-score indicate strong overall classification performance. The precision and recall scores highlight the model's effectiveness in correctly identifying damaged cases while limiting false positives. The high AUC demonstrates the DIP's strong capability to discriminate between damaged and undamaged instances. A decision threshold of 0.444 was selected to balance precision and recall optimally, based on precision-recall trade-off analysis.

\subsubsection{Damage Outcome Generator Evaluation}
\label{sec:model_eval_dog}

The Damage Outcome Generator is evaluated under both reconstruction and generative settings.

In reconstruction mode, the model achieves a Mean Absolute Error (MAE) of 13.97 HP (13.97\% of the total damage scale) and an $R^2$ score of 0.687. The hit group classification accuracy reaches 96.6\%. Both results indicate effective reconstruction of damage outcomes. We attribute the remaining prediction errors primarily to the lack of temporal context and the assumption of independent pairwise interactions (see Discussion).

\begin{figure}[!htbp]
\centering
\includegraphics[width=0.9\linewidth]{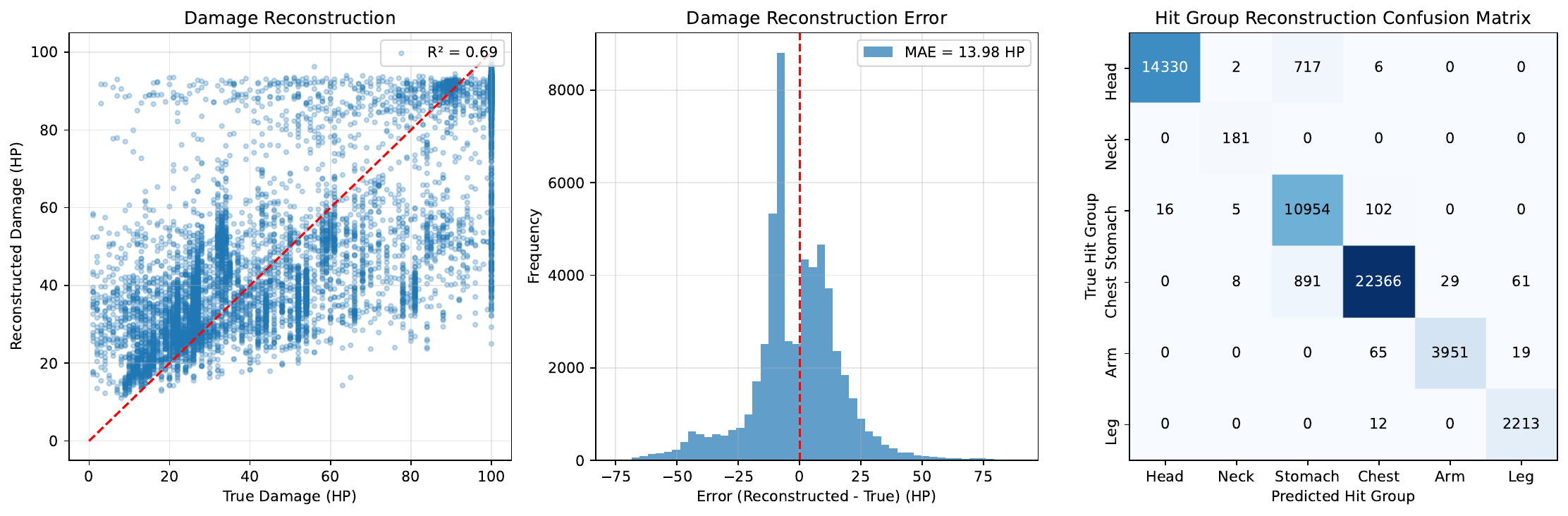}
\caption{DOG reconstruction evaluation. Left: Actual vs. reconstructed damage ($R^2 = 0.687$). Center: Reconstruction error histogram. Right: Hit group confusion matrix (accuracy = 96.6\%).}
\label{fig:damage_generation_eval}
\end{figure}

\begin{figure}[!htbp]
\centering
\includegraphics[width=0.8\linewidth]{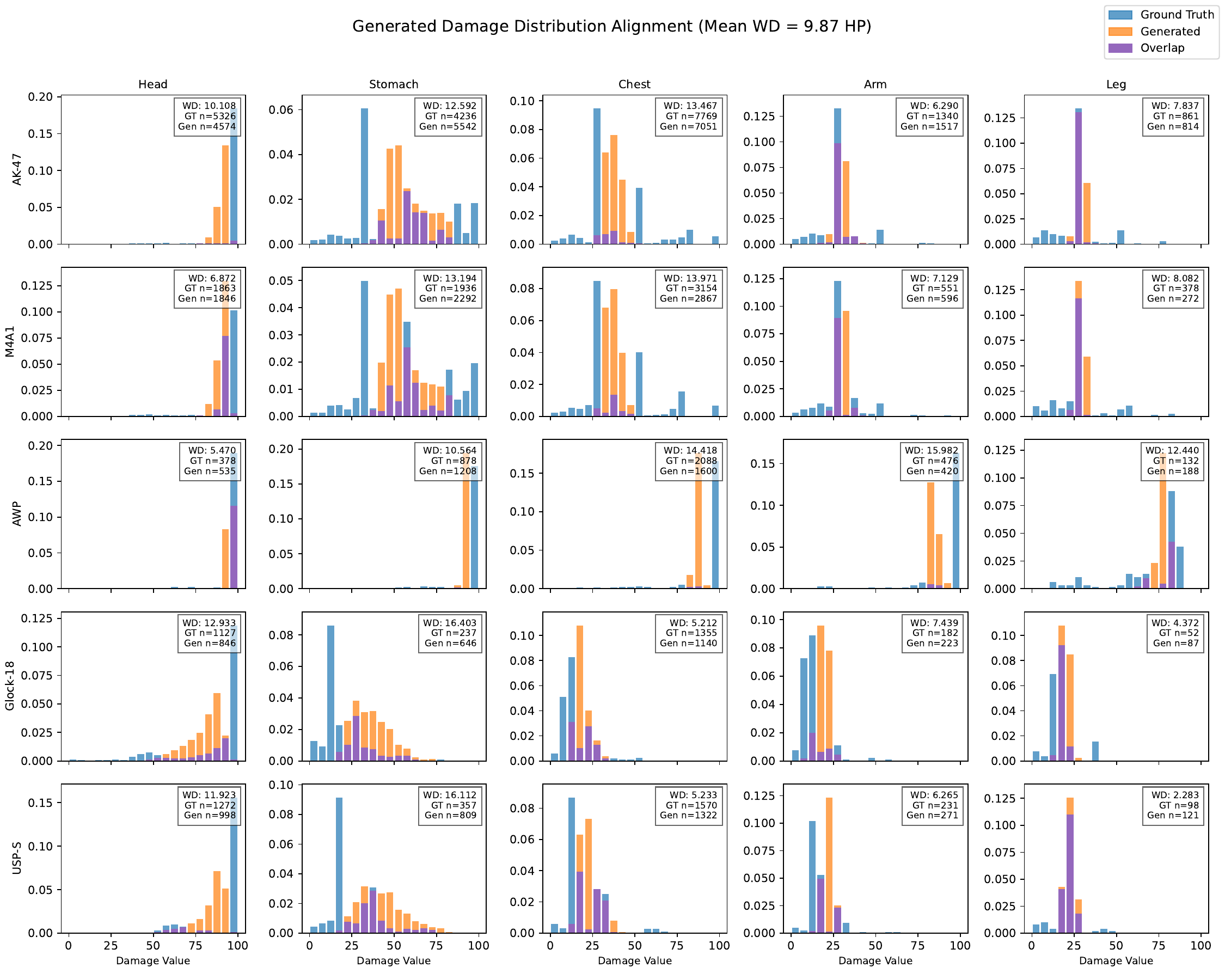}
\caption{Distributional alignment between generated and true damage patterns across the 5 most common weapons and hit groups. The mean Wasserstein Distance (WD) between predicted and true distributions is 9.87 Health Point (HP)s (9.87\% of the total damage scale).}
\label{fig:damage_dist_eval}
\end{figure}

In generative mode, the model's performance is assessed through distributional alignment metrics. Figure~\ref{fig:damage_dist_eval} compares the generated and actual damage distributions across weapon types and hit locations, with mean Wasserstein Distance (WD) 9.87 HP (9.87\% of the total damage scale), and confirming that the model generates statistically accurate outcomes.

\subsubsection{Latent Space Analysis}
\label{sec:latent}

\begin{figure}[!htbp]
\centering
\includegraphics[width=0.63\linewidth]{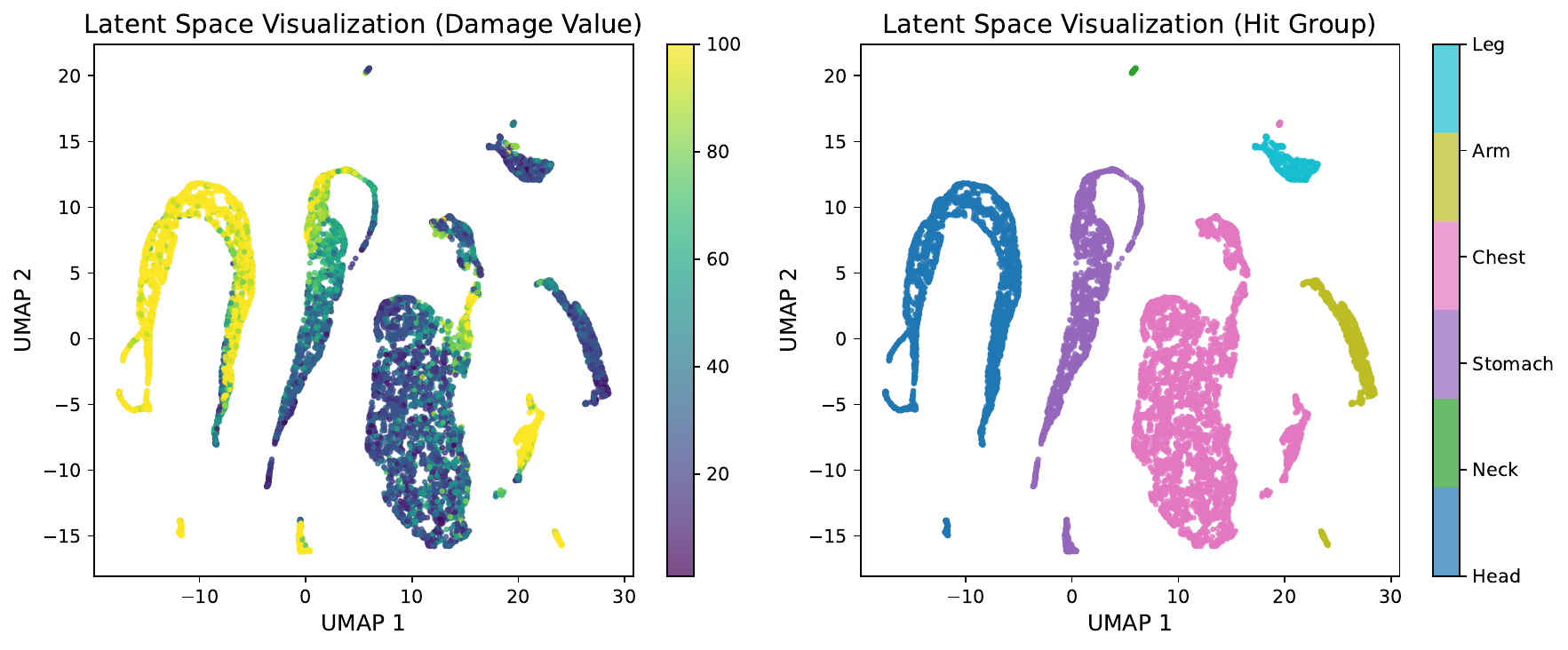}
\caption{UMAP plot of DOG's latent space. Left: Colored by damage value. Right: Colored by hit-group.}
\label{fig:latent_space}
\end{figure}

We analyze the latent space learned by the DOG model using Unified Manifold Approximation and Projection (UMAP)~\shortcite{mcinnes2018umap} to project latent vectors into two dimensions. Figure~\ref{fig:latent_space} shows the projection colored by damage values (left) and hit groups (right). The damage view reveals smooth gradients from low to high values, while the hit group view shows clearly separated clusters for different body parts. This indicates that the representation jointly captures a continuous encoding of damage severity and a categorical encoding of hit group.

\vspace{-0.4cm}
\subsection{Human Trajectory Replay}

To assess the realism of movement in our simulation environment, we evaluate how well DECOY replicates human player behavior by replaying professional CS:GO trajectories within the simulator.

We begin by converting movement data from the ESTA dataset into waypoint trajectories by mapping each timestamp to its nearest waypoint. If a step skips multiple waypoints, we interpolate the missing points using shortest path algorithms. This results in a waypoint trajectory that can be translated into a discrete action sequence using only the eight compass directions. The resulting sequence is replayed in the simulator with all game mechanics enabled, including shooting and damage via DIP-DOG, bomb interactions, and final outcomes.

To compare the replayed trajectories with the original human data, we sample simulator outputs at the same frequency as the original trajectory (2 FPS) and normalize all trajectories to the maximum sequence length via interpolation. We then evaluate the spatial and temporal similarity between the original and waypoint trajectories using various metrics, alongside comparisons of health point distribution over time and final game outcomes.

\subsubsection{Spatiotemporal Alignment of Movement Trajectories}

\begin{figure}[!htbp]
\centering
\includegraphics[width=0.82\linewidth]{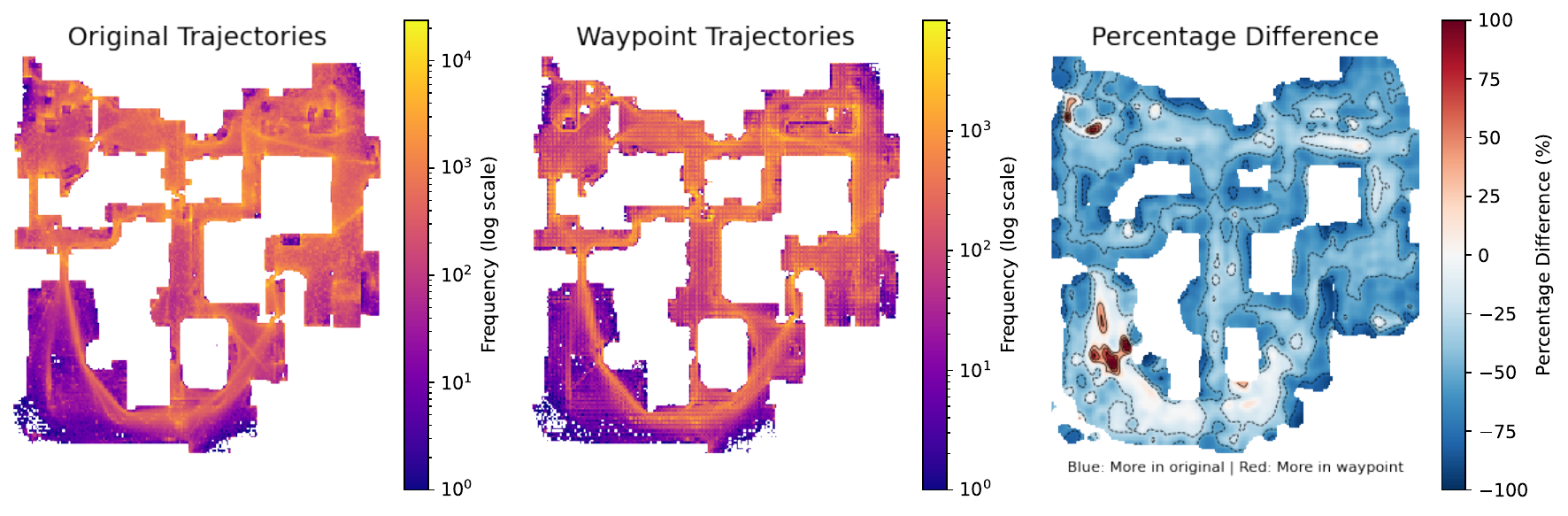}
\caption{Distribution comparison between the original player trajectory and the replayed waypoint trajectory. Left: Heatmap distribution of original trajectories — brighter colors indicate more frequently visited routes. Middle: Heatmap of waypoint trajectories. Right: Percentage difference contour plots.}
\label{fig:trajectory_comparison_heatmap}
\end{figure}

We evaluate trajectory fidelity using a set of geometric and temporal similarity metrics: Dynamic Time Warping (DTW) \shortcite{salvador2007toward}, Euclidean distance, Root Mean Squared Error (RMSE), and Fréchet distance. To isolate the effect of external factors such as damage and player death, we evaluate movement trajectories with shooting and bomb actions disabled. Each metric is computed per player across all rounds and aggregated by team (T vs. CT) as well as overall, with results reported in Table~\ref{tab:trajectory_alignment}.

\begin{table}[ht]
\centering
\caption{Trajectory alignment metrics comparing replayed (waypoint) and original human trajectories. Values are reported as mean $\pm$ standard deviation. Lower is better.}
\label{tab:trajectory_alignment}
\resizebox{0.6\textwidth}{!}{%
\begin{tabular}{lccc}
\toprule
\textbf{Metric} & \textbf{T Side} & \textbf{CT Side} & \textbf{Overall} \\
\midrule
DTW            & $0.873 \pm 1.868$ & $1.332 \pm 2.263$ & $1.103 \pm 2.088$ \\
Euclidean      & $0.420 \pm 0.237$ & $0.467 \pm 0.252$ & $0.443 \pm 0.246$ \\
RMSE           & $5.155 \pm 2.911$ & $5.720 \pm 3.091$ & $5.437 \pm 3.016$ \\
Fréchet 2D     & $4.030 \pm 8.599$ & $6.690 \pm 9.910$ & $5.360 \pm 9.372$ \\
\bottomrule
\end{tabular}
}
\end{table}

The waypoint-based trajectories generally align with the original human trajectories with high spatial and temporal fidelity. Notably, alignment is tighter on the T side than the CT side, possibly because bombsites A and B—closer to the CT spawn—are more cluttered with obstacles, making accurate replay more difficult. The average Euclidean distance is 0.443 meters (with waypoint spacing at 0.7 meters), meaning agents stay within about two-thirds of a waypoint from the intended path, indicating strong spatial accuracy. The higher DTW value compared to Euclidean distance suggests minor temporal misalignments, though the overall trajectory shapes remain preserved. The elevated Fréchet distance (5.36) and RMSE indicate occasional detours, likely due to limited waypoint coverage in less-traveled areas.

We further visualize both the original and waypoint trajectories as heatmaps in Figure~\ref{fig:trajectory_comparison_heatmap}. Brighter regions indicate more commonly visited paths, and the heatmaps show strong overall alignment between the two. Due to discretization, the waypoint trajectory appears pixelated. We also plot the percentage difference between the two as a contour map, revealing that misalignments tend to occur near walls and obstacles—especially in areas requiring jumps. These deviations may result from agents accidentally falling off ledges or taking unintended detours.

\subsubsection{Health Points and Game Results}
\begin{figure}[!htbp]
\centering
\includegraphics[width=0.8\linewidth]{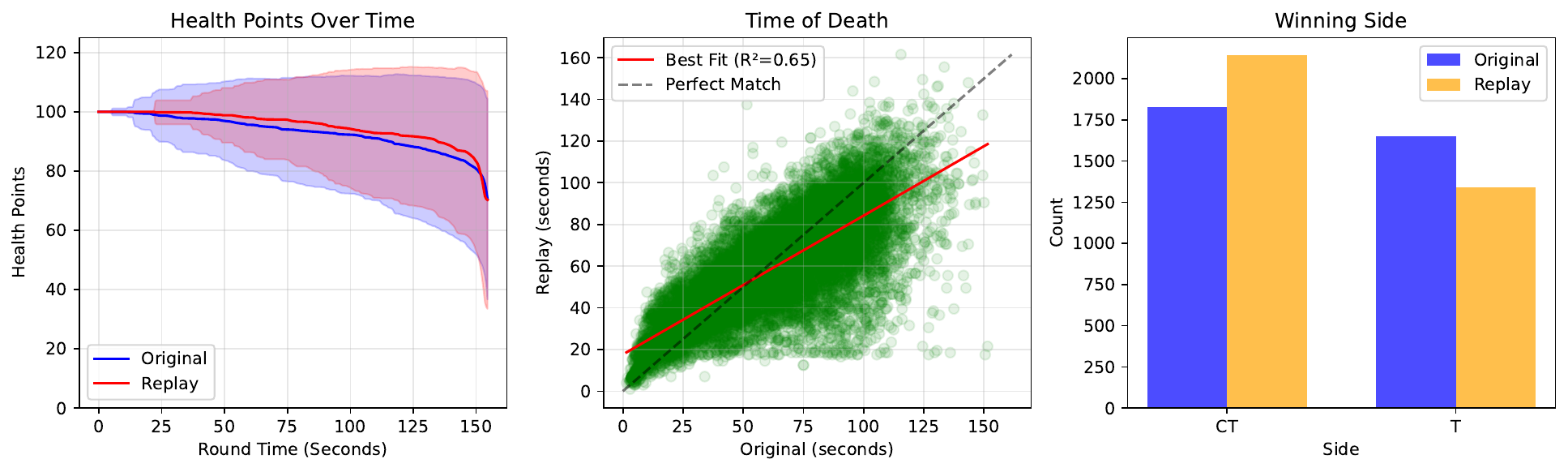}
\caption{Human replay results. Left: Health point distribution over time (mean $\pm$ std). Middle: Correlation between time of death and other factors. Right: Final game outcome distribution.}
\label{fig:trajectory_analysis}
\end{figure}

To evaluate how accurately DECOY reproduces gameplay outcomes, we analyze health point trajectories, death timings, and final match results. A combination of damage event models and line-of-sight detection is used to simulate realistic combat interactions. Bomb planting and defusing are automated: the bomb is planted when the carrier reaches the bombsite, and defusing is triggered when a CT agent reaches the bomb. As shown in Figure~\ref{fig:trajectory_analysis}, DECOY achieves strong alignment with the original game in terms of health progression (HP Correlation: 0.961; HP RMSE: 2.35) and reasonable agreement in death timings (Death Time Correlation: 0.809; MAE: 12.8 seconds). The final match outcome aligns with the original 91.0\% of the time. However, the underlying reasons for team victories can differ, largely due to the current DIP-DOG model’s assumptions—specifically, the lack of temporal dependency modeling and the use of pairwise conditional independence. These factors further compound to inaccurate bombing actions (see Discussion), where future work is needed to improve fidelity.

\section{Discussion}

While DECOY provides a fast and tactically grounded simulation for modeling team-based combat in CS:GO at a strategic level, accurately predicting and generating every aspect of the game using only movement remains challenging. A notable issue is the compounding error effect involving movement, shooting damage, and final game outcomes. Movement replay errors can cause agents to drift from their intended positions and timings, which leads to incorrect conflict timings and inaccurate damage calculations. These inaccuracies further cascade into distorted outcomes, particularly around bomb-related actions, as agents may not expire at the correct moments. However, achieving perfect replication of exact CS:GO game outcomes is beyond the scope of this paper.

Several limitations also exist in our current implementation and modeling. DECOY models each round independently, neglecting longer-term factors like team economy and multi-round strategic planning—both essential for realistic agent behavior. For example, players may choose to hide and save weapons for future rounds when at a disadvantage, a tactic not currently represented. Additionally, simulation performance slows as agent count increases. Although physics settings are set to high during replay to match the original game, users can improve speed by lowering physics update frequency and increasing agent speed. The current simulation also excludes utility mechanics such as grenades, flashbangs, and smoke. Moreover, the DIP and DOG models do not incorporate temporal dependencies, and assume conditional independence in pairwise interactions. This limitation restricts their ability to fully capture the complex joint dynamics and temporal relationships inherent in multi-agent engagements. Empirical evaluation indicates that this lack of temporal dependency often results in premature agent deaths, leading to further inaccuracies in bomb-related actions during replays. Future work will address these limitations by incorporating more expressive temporal modeling methods, such as Graph Neural Networks and Transformers.

Finally, while DECOY is specifically grounded in the structure and mechanics of CS:GO, our broader aim is to support research into tactical and strategic decision-making in real-world scenarios. Recent advances in 3D scene reconstruction, wearable tracking devices, and video-based motion capture technologies increasingly enable realistic environment reconstruction with accurate human movement data. By abstracting such physical environments into navigable waypoint networks and modeling high-level decision events, the simulation framework developed within DECOY holds potential to extend to diverse real-world tactical and strategic domains with the integration of generative models.

\vspace{-0.3cm}
\section{Conclusion}

\vspace{-0.2cm}
We presented DECOY, a fast, data-driven multi-agent simulation environment inspired by tactical shooter gameplay. By combining waypoint-based navigation with a modular two-stage generative model for damage prediction, DECOY enables efficient and high-fidelity simulation of complex multi-agent engagements. Its alignment with human trajectory data and incorporation of stochastic combat modeling allow it to capture both strategic behavior and the uncertainty inherent in real-world scenarios. DECOY offers a flexible and extensible platform for advancing research in multi-agent tactical decision-making.

\vspace{-0.3cm}
\section*{Acknowledgments} 
\vspace{-0.1cm}
The authors acknowledge the use of Large Language Models for assistance with proofreading and grammar checking. All content was reviewed, edited, and approved by the human authors, who take full responsibility for the final manuscript. The project or effort depicted was or is sponsored by the U.S. Army Combat Capabilities Development Command -- Soldier Centers under contract number W912CG-24-D-0001. The content of the information does not necessarily reflect the position or the policy of the Government, and no official endorsement should be inferred.

\footnotesize

\bibliographystyle{wsc}

\vspace{-0.3cm}
\bibliography{ref.bib}

\begin{thebibliography}{}

\bibitem[\protect\citeauthoryear{Baker, Kanitscheider, Markov, Wu, Powell,
  McGrew, and Mordatch}{Baker et~al.}{2019}]{baker2019emergent}
Baker, B., I.~Kanitscheider, T.~Markov, Y.~Wu, G.~Powell, B.~McGrew {\em
  et~al}. 2019.
\newblock ``Emergent Tool Use from Multi-Agent Autocurricula''.
\newblock In {\em Proceedings of the International Conference on Learning
  Representations}.
\newblock New Orleans, Louisiana, United States.
\newblock May 6--9, 2019.

\bibitem[\protect\citeauthoryear{Berner, Brockman, Chan, Cheung, D{\k{e}}biak,
  Dennison, Farhi, Fischer, Hashme, Hesse, et~al.}{Berner
  et~al.}{2019}]{berner2019dota}
Berner, C., G.~Brockman, B.~Chan, V.~Cheung, P.~D{\k{e}}biak, C.~Dennison,
  {\em et~al}. 2019.
\newblock ``Dota 2 with Large Scale Deep Reinforcement Learning''.
\newblock {\em arXiv preprint arXiv:1912.06680\/}.


\bibitem[\protect\citeauthoryear{{Carnegie Mellon Entertainment Technology
  Center and Walt Disney Imagineering}}{{Carnegie Mellon Entertainment
  Technology Center and Walt Disney Imagineering}}{2024}]{panda3d}
{Carnegie Mellon Entertainment Technology Center and Walt Disney Imagineering}
  2024.
\newblock ``Panda3D''.
\newblock {\em https://www.panda3d.org\/}.
\newblock Version 1.10.14, accessed 23rd March.


\bibitem[\protect\citeauthoryear{Community}{Community}{2025}]{bspsource}
Community, V.~D. 2025.
\newblock ``BSPSource''.
\newblock {\em https://developer.valvesoftware.com/wiki/BSPSource\/}.
\newblock accessed 23rd March.


\bibitem[\protect\citeauthoryear{Durst, Xie, Sarukkai, Shacklett, Frosio,
  Tessler, Kim, Taylor, Bernstein, Choudhury, et~al.}{Durst
  et~al.}{2024}]{durst2024csgo}
Durst, D., F.~Xie, V.~Sarukkai, B.~Shacklett, I.~Frosio, C.~Tessler,  {\em
  et~al}. 2024.
\newblock ``Learning to Move Like Professional Counter-Strike Players''.
\newblock In {\em Computer Graphics Forum}, Volume~43,  e15173.
\newblock Wiley Online Library.

\bibitem[\protect\citeauthoryear{Garrido, Assran, Ballas, Bardes, Najman, and
  LeCun}{Garrido et~al.}{2024}]{garrido2024learning}
Garrido, Q., M.~Assran, N.~Ballas, A.~Bardes, L.~Najman, and Y.~LeCun. 2024.
\newblock ``Learning and Leveraging World Models in Visual Representation
  Learning''.
\newblock {\em arXiv preprint arXiv:2403.00504\/}.


\bibitem[\protect\citeauthoryear{Ha and Schmidhuber}{Ha and
  Schmidhuber}{2018}]{ha2018world}
Ha, D., and J.~Schmidhuber. 2018.
\newblock ``World Models''.
\newblock {\em arXiv preprint arXiv:1803.10122\/}.


\bibitem[\protect\citeauthoryear{Hafner, Pasukonis, Ba, and Lillicrap}{Hafner
  et~al.}{2023}]{hafner2023mastering}
Hafner, D., J.~Pasukonis, J.~Ba, and T.~Lillicrap. 2023.
\newblock ``Mastering Diverse Domains through World Models''.
\newblock {\em arXiv preprint arXiv:2301.04104\/}.


\bibitem[\protect\citeauthoryear{Kaplan, McCandlish, Henighan, Brown, Chess,
  Child, Gray, Radford, Wu, and Amodei}{Kaplan
  et~al.}{2020}]{kaplan2020scaling}
Kaplan, J., S.~McCandlish, T.~Henighan, T.~B. Brown, B.~Chess, R.~Child,  {\em
  et~al}. 2020.
\newblock ``Scaling Laws for Neural Language Models''.
\newblock {\em arXiv preprint arXiv:2001.08361\/}.


\bibitem[\protect\citeauthoryear{Kingma and Welling}{Kingma and
  Welling}{2013}]{kingma2013auto}
Kingma, D.~P., and M.~Welling. 2013.
\newblock ``Auto-Encoding Variational Bayes''.
\newblock {\em arXiv preprint arXiv:1312.6114\/}.


\bibitem[\protect\citeauthoryear{Koresh, Ustun, Kumar, and Aris}{Koresh
  et~al.}{2024}]{koresh2024improving}
Koresh, C., V.~Ustun, R.~Kumar, and T.~Aris. 2024, May.
\newblock ``Improving Reinforcement Learning Experiments in Unity through
  Waypoint Utilization''.
\newblock In {\em The International FLAIRS Conference Proceedings}, Volume~37.

\bibitem[\protect\citeauthoryear{Laskin, Wang, Oh, Parisotto, Spencer,
  Steigerwald, et~al.}{Laskin et~al.}{2022}]{laskin2022context}
Laskin, M., L.~Wang, J.~Oh, E.~Parisotto, S.~Spencer, R.~Steigerwald {\em
  et~al}. 2022.
\newblock ``In-Context Reinforcement Learning with Algorithm Distillation''.
\newblock {\em arXiv preprint arXiv:2210.14215\/}.


\bibitem[\protect\citeauthoryear{Li, Wang, Wang, Jin, Li, Zeng, and Yang}{Li
  et~al.}{2024}]{li2024open}
Li, J., Q.~Wang, Y.~Wang, X.~Jin, Y.~Li, W.~Zeng {\em et~al}. 2024.
\newblock ``Open-World Reinforcement Learning over Long Short-Term
  Imagination''.
\newblock {\em arXiv preprint arXiv:2410.03618\/}.


\bibitem[\protect\citeauthoryear{Mathieu, Ozair, Srinivasan, Gulcehre, Zhang,
  Jiang, Le~Paine, Powell, {\.Z}o{\l}na, Schrittwieser, et~al.}{Mathieu
  et~al.}{2023}]{mathieu2023alphastar}
Mathieu, M., S.~Ozair, S.~Srinivasan, C.~Gulcehre, S.~Zhang, R.~Jiang,  {\em
  et~al}. 2023.
\newblock ``AlphaStar Unplugged: Large-Scale Offline Reinforcement Learning''.
\newblock {\em arXiv preprint arXiv:2308.03526\/}.


\bibitem[\protect\citeauthoryear{McInnes, Healy, Saul, and
  Gro{\ss}berger}{McInnes et~al.}{2018}]{mcinnes2018umap}
McInnes, L., J.~Healy, N.~Saul, and L.~Gro{\ss}berger. 2018.
\newblock ``UMAP: Uniform Manifold Approximation and Projection''.
\newblock {\em Journal of Open Source Software\/}~3(29):861.


\bibitem[\protect\citeauthoryear{Neumann and Gros}{Neumann and
  Gros}{2022}]{neumann2022scaling}
Neumann, O., and C.~Gros. 2022.
\newblock ``Scaling Laws for a Multi-Agent Reinforcement Learning Model''.
\newblock {\em arXiv preprint arXiv:2210.00849\/}.


\bibitem[\protect\citeauthoryear{Nikulin, Zisman, Zemtsov, Sinii, Kurenkov, and
  Kolesnikov}{Nikulin et~al.}{2024}]{nikulin2024xland}
Nikulin, A., I.~Zisman, A.~Zemtsov, V.~Sinii, V.~Kurenkov, and S.~Kolesnikov.
  2024.
\newblock ``XLand-100B: A Large-Scale Multi-Task Dataset for In-Context
  Reinforcement Learning''.
\newblock {\em arXiv preprint arXiv:2406.08973\/}.


\bibitem[\protect\citeauthoryear{Obando-Ceron, Sokar, Willi, Lyle, Farebrother,
  Foerster, Dziugaite, Precup, and Castro}{Obando-Ceron
  et~al.}{2024}]{obando2024mixtures}
Obando-Ceron, J., G.~Sokar, T.~Willi, C.~Lyle, J.~Farebrother, J.~Foerster,
  {\em et~al}. 2024.
\newblock ``Mixtures of Experts Unlock Parameter Scaling for Deep {RL}''.
\newblock {\em arXiv preprint arXiv:2402.08609\/}.


\bibitem[\protect\citeauthoryear{Omidshafiei, Hennes, Garnelo, Wang, Recasens,
  Tarassov, Yang, Elie, Connor, Muller, et~al.}{Omidshafiei
  et~al.}{2022}]{omidshafiei2022multiagent}
Omidshafiei, S., D.~Hennes, M.~Garnelo, Z.~Wang, A.~Recasens, E.~Tarassov,
  {\em et~al}. 2022.
\newblock ``Multiagent Off-Screen Behavior Prediction in Football''.
\newblock {\em Scientific Reports\/}~12(1):8638.


\bibitem[\protect\citeauthoryear{O'Neill, Rehman, Maddukuri, Gupta, Padalkar,
  Lee, et~al.}{O'Neill et~al.}{2024}]{o2024open}
O'Neill, A., A.~Rehman, A.~Maddukuri, A.~Gupta, A.~Padalkar, A.~Lee {\em
  et~al}. 2024.
\newblock ``Open {X-Embodiment}: Robotic Learning Datasets and {RT-X} Models:
  Open {X-Embodiment} Collaboration 0''.
\newblock In {\em Proceedings of the 2024 IEEE International Conference on
  Robotics and Automation},  6892--6903.
\newblock Yokohama, Japan: IEEE.
\newblock May 13–17, 2024.

\bibitem[\protect\citeauthoryear{{Open Ended Learning Team}, Stooke, Mahajan,
  Barros, Deck, Bauer, et~al.}{{Open Ended Learning Team}
  et~al.}{2021}]{team2021open}
{Open Ended Learning Team}, A.~Stooke, A.~Mahajan, C.~Barros, C.~Deck, J.~Bauer
  {\em et~al}. 2021.
\newblock ``Open-Ended Learning Leads to Generally Capable Agents''.
\newblock {\em arXiv preprint arXiv:2107.12808\/}.


\bibitem[\protect\citeauthoryear{Reed, Zolna, Parisotto, Colmenarejo, Novikov,
  Barth-Maron, et~al.}{Reed et~al.}{2022}]{reed2022generalist}
Reed, S., K.~Zolna, E.~Parisotto, S.~G. Colmenarejo, A.~Novikov, G.~Barth-Maron
  {\em et~al}. 2022.
\newblock ``A Generalist Agent''.
\newblock {\em arXiv preprint arXiv:2205.06175\/}.


\bibitem[\protect\citeauthoryear{Salvador and Chan}{Salvador and
  Chan}{2007}]{salvador2007toward}
Salvador, S., and P.~Chan. 2007.
\newblock ``Toward Accurate Dynamic Time Warping in Linear Time and Space''.
\newblock {\em Intelligent Data Analysis\/}~11(5):561--580.


\bibitem[\protect\citeauthoryear{Socolow and Jolly}{Socolow and
  Jolly}{2017}]{socolow2017game}
Socolow, B., and I.~Jolly. 2017.
\newblock ``Game-Changing Wearable Devices That Collect Athlete Data Raise Data
  Ownership Issues''.
\newblock {\em World Sports Advocate\/}~15(7):15--17.


\bibitem[\protect\citeauthoryear{Sohn, Lee, and Yan}{Sohn
  et~al.}{2015}]{sohn2015learning}
Sohn, K., H.~Lee, and X.~Yan. 2015.
\newblock ``Learning Structured Output Representation using Deep Conditional
  Generative Models''.
\newblock In {\em Advances in Neural Information Processing Systems}, edited
  by\ C.~Cortes, N.~Lawrence, D.~Lee, M.~Sugiyama, and R.~Garnett, Volume~28:
  Curran Associates, Inc.

\bibitem[\protect\citeauthoryear{Terry, Black, Grammel, Jayakumar, Hari,
  Sullivan, et~al.}{Terry et~al.}{2021}]{terry2021pettingzoo}
Terry, J., B.~Black, N.~Grammel, M.~Jayakumar, A.~Hari, R.~Sullivan {\em
  et~al}. 2021.
\newblock ``PettingZoo: Gym for Multi-Agent Reinforcement Learning''.
\newblock {\em Advances in Neural Information Processing
  Systems\/}~34:15032--15043.


\bibitem[\protect\citeauthoryear{Tuyls, Omidshafiei, Muller, Wang, Connor,
  Hennes, Graham, Spearman, Waskett, Steel, et~al.}{Tuyls
  et~al.}{2021}]{tuyls2021game}
Tuyls, K., S.~Omidshafiei, P.~Muller, Z.~Wang, J.~Connor, D.~Hennes,  {\em
  et~al}. 2021.
\newblock ``Game Plan: What {AI} Can Do for Football, and What Football Can Do
  for {AI}''.
\newblock {\em Journal of Artificial Intelligence Research\/}~71:41--88.


\bibitem[\protect\citeauthoryear{Ustun, Hans, Kumar, and Wang}{Ustun
  et~al.}{2024}]{ustun2024geospecific}
Ustun, V., S.~Hans, R.~Kumar, and Y.~Wang. 2024.
\newblock ``Abstracting Geo-Specific Terrains to Scale Up Reinforcement
  Learning''.
\newblock In {\em Proceedings of the Interservice/Industry Training, Simulation
  and Education Conference (I/ITSEC) 2024}.
\newblock Orlando, Florida, United States: National Training and Simulation
  Association (NTSA).

\bibitem[\protect\citeauthoryear{Vinyals, Babuschkin, Czarnecki, Mathieu,
  Dudzik, Chung, Choi, Powell, Ewalds, Georgiev, et~al.}{Vinyals
  et~al.}{2019}]{vinyals2019grandmaster}
Vinyals, O., I.~Babuschkin, W.~M. Czarnecki, M.~Mathieu, A.~Dudzik, J.~Chung,
  {\em et~al}. 2019.
\newblock ``Grandmaster Level in {StarCraft II} Using Multi-Agent Reinforcement
  Learning''.
\newblock {\em Nature\/}~575(7782):350--354.


\bibitem[\protect\citeauthoryear{Wang, Veli{\v{c}}kovi{\'c}, Hennes,
  Tomas{\v{s}}ev, Prince, Kaisers, Bachrach, Elie, Wenliang, Piccinini,
  et~al.}{Wang et~al.}{2024}]{wang2024tacticai}
Wang, Z., P.~Veli{\v{c}}kovi{\'c}, D.~Hennes, N.~Tomas{\v{s}}ev, L.~Prince,
  M.~Kaisers,  {\em et~al}. 2024.
\newblock ``Tactic{AI}: An {AI} Assistant for Football Tactics''.
\newblock {\em Nature Communications\/}~15(1):1906.


\bibitem[\protect\citeauthoryear{Xenopoulos}{Xenopoulos}{2023}]{xenopoulos2023data}
Xenopoulos, P. 2023.
\newblock {\em Data Mining for Esports}.
\newblock Ph.\ D. thesis, New York University Tandon School of Engineering.

\bibitem[\protect\citeauthoryear{Xenopoulos and Silva}{Xenopoulos and
  Silva}{2022}]{xenopoulos2022esta}
Xenopoulos, P., and C.~Silva. 2022.
\newblock ``ESTA: An Esports Trajectory and Action Dataset''.
\newblock {\em arXiv preprint arXiv:2209.09861\/}.


\end{thebibliography}

\vspace{-0.4cm}
\section*{AUTHOR BIOGRAPHIES}

\noindent {\bf \MakeUppercase{Yunzhe Wang}} is a Ph.D. student at the University of Southern California. His research interests include Multi-Agent Learning, Sequential Decision-making, Reinforcement Learning, and Large Language Models. Email: \href{mailto:yunzhewa@usc.edu}{yunzhewa@usc.edu}

\vspace{0.5em}

\noindent {\bf \MakeUppercase{Volkan Ustun}} is the Associate Director of the Human-Inspired Adaptive Teaming Systems Group at the USC Institute for Creative Technologies. His research augments Multi-agent Reinforcement Learning (MARL) models, drawing inspiration from operations research, human judgment and decision-making, game theory, graph theory, and cognitive architectures to better address the challenges of developing behavior models for synthetic characters, mainly in military training simulations. Email: \href{mailto:ustun@ict.usc.edu}{ustun@ict.usc.edu}

\vspace{0.5em}

\noindent {\bf \MakeUppercase{Chris McGroarty}} is the Chief Engineer for Advanced Simulation at the US Army Combat Capabilities Development Command, Soldier Center, Simulation and Training Technology Center (DEVCOM SC STTC). His research interests include distributed simulation, simulation architectures, applications of Artificial Intelligence Technologies to simulation, novel computing architectures, innovative methods for user-simulation interaction, methodologies for making simulation more accessible by non-simulation experts, future simulation frameworks, and the application of videogame industry technologies. Email: \href {mailto:christopher.j.mcgroarty.civ@army.mil}{christopher.j.mcgroarty.civ@army.mil}

\end{document}